\documentclass[conference]{IEEEtran}
\usepackage{cite}
\usepackage{amsmath,amssymb,amsfonts}

\usepackage{algorithm}
\usepackage{algcompatible}
\usepackage{algorithmicx}
\usepackage[noend]{algpseudocode}

\usepackage{graphicx}
\usepackage{textcomp}
\usepackage{xcolor}
\def\BibTeX{{\rm B\kern-.05em{\sc i\kern-.025em b}\kern-.08em
    T\kern-.1667em\lower.7ex\hbox{E}\kern-.125emX}}
    
\usepackage{subcaption}
\usepackage{comment}
\usepackage{hyperref} 
\hypersetup{
    colorlinks=true,
    linkcolor=blue,
    filecolor=blue,      
    urlcolor=blue,
    citecolor=blue
}

\usepackage{todonotes}
\usepackage{multirow}
\graphicspath{{figures/}}
    \DeclareGraphicsExtensions{.pdf,.eps,.png,.jpg}
\usepackage{bm}	
\usepackage{booktabs}
\usepackage[noabbrev,capitalise]{cleveref} 
\crefname{equation}{}{}	
\usepackage{tikz}
\usepackage{upgreek, tikz, standalone}
\usepackage[utf8]{inputenc}
\usepackage[acronym]{glossaries}
\newcommand*{\NmarginT}{\ensuremath{T}}
\newcommand*{\NspecificityS}{\ensuremath{s}}

\usepackage[nolist]{acronym}
\usepackage[english]{babel}

\begin{acronym}[DRAM]
    \acro{AI}{Artificial Intelligence}
    \acrodefindefinite{AI}{an}{an}
    \acro{NN}{Neural Network}
    \acro{DL}{Deep Learning}
    \acro{ML}{Machine Learning}
    \acro{TM}{Tsetlin Machine}
    \acro{NLP}{Natural Language Processing}
    \acro{TM-AE}{Tsetlin Machine Auto-Encoder}
    \acro{RbE}{Reasoning by Elimination}
    \acro{TW}{Target Word}
    \acro{CoTM}{Coalesced Tsetlin Machine}
    \acrodefindefinite{ML}{an}{a}
    \acro{TA}{Tsetlin automaton}
    \acrodefplural{TA}{Tsetlin automata}
    \acro{TAT}{Tsetlin automaton team}
\end{acronym}

\usepackage{algpseudocode}

\algnewcommand{\LineComment}[1]{\State \(\triangleright\) #1}

\begin{document}
\title{Exploring State Space and Reasoning by Elimination in Tsetlin Machines}
\author{
\IEEEauthorblockN{Ahmed K. Kadhim}
\IEEEauthorblockA{\textit{Department of ICT} \\
\textit{University of Agder}\\
Grimstad, Norway \\
ahmed.k.kadhim@uia.no}
\and
\IEEEauthorblockN{Ole-Christoffer Granmo}
\IEEEauthorblockA{\textit{Department of ICT} \\
\textit{University of Agder}\\
Grimstad, Norway \\
ole.granmo@uia.no}
\and
\IEEEauthorblockN{Lei Jiao}
\IEEEauthorblockA{\textit{Department of ICT} \\
\textit{University of Agder}\\
Grimstad, Norway \\
lei.jiao@uia.no}
\and
\IEEEauthorblockN{Rishad Shafik}
\IEEEauthorblockA{\textit{School of Engineering} \\
\textit{Newcastle University}\\
Newcastle upon Tyne, UK \\
rishad.shafik@newcastle.ac.uk}
}

\maketitle

\begin{abstract}
The \ac{TM} has gained significant attention in \ac{ML}. By employing logical fundamentals, it facilitates pattern learning and representation, offering an alternative approach for developing comprehensible \ac{AI} with a specific focus on pattern classification in the form of conjunctive clauses. In the domain of \ac{NLP}, \ac{TM} is utilised to construct word embedding and describe target words using clauses. To enhance the descriptive capacity of these clauses, we study the concept of \ac{RbE} in clauses' formulation, which involves incorporating feature negations to provide a more comprehensive representation.
In more detail, this paper employs the \ac{TM-AE} architecture to generate dense word vectors, aiming at capturing contextual information by extracting feature-dense vectors for a given vocabulary. 
Thereafter, the principle of \ac{RbE} is explored to improve descriptivity and optimise the performance of the \ac{TM}.  
Specifically, the specificity parameter \NspecificityS{} and the voting margin parameter \NmarginT{} are leveraged to regulate feature distribution in the state space, resulting in a dense representation of information for each clause. In addition, we investigate the state spaces of \ac{TM-AE}, especially for the forgotten/excluded features.  Empirical investigations on artificially generated data, the IMDB dataset, and the 20 Newsgroups dataset showcase the robustness of the \ac{TM}, with accuracy reaching 90.62\% for the IMDB.
\end{abstract}
\section{Introduction}
\label{sec:intro}


In natural language processing applications, \ac{TM} extracts context information for an input by identifying its active features and performing an AND operation to generate conjunctive clauses.  The process of building these clauses involves memorising or forgetting features within the clause features' space in memory~\cite{granmo2018tsetlin}. 

In a recent study, the \ac{TM}-based auto-encoder for self-supervised embedding has been implemented with human-interpretable contextual representations from a given dataset~\cite{bhattarai2023tsetlin}.
The model demonstrated promising performance comparable to Word2Vec~\cite{mikolov2013efficient}, FastText~\cite{bojanowski2017enriching}, and GLoVe~\cite{pennington2014glove} across multiple datasets~\cite{goldberg2014word2vec, mikolov2013distributed}.
The main focus of~\cite{bhattarai2023tsetlin} is on word embedding for a group of input words, rather than obtaining context information for an individual word.  

It has been observed in~\cite{yadav2021human} that literals with negated features are common in a clause, which coincides with the concept of \ac{RbE}. In addition, using negated reasoning in the \ac{TM}, the work in~\cite{yadav2022robust} proposed the augmentation of clause robustness by decreasing the specificity factor \NspecificityS{}, which has displayed promising improvements in the accuracy of downstream classification tasks.   
While empirical observations support this approach, there remains a dearth of explanations regarding the precise impact of the hyperparameter modification on the \ac{TM}'s structure. Although the transparent nature of the \ac{TM} model renders it mathematically comprehensible, further investigation is warranted to ascertain why \NspecificityS{} was selected and whether alternatives exist, which can possibly provide superior results. This necessitates obtaining clauses that possess specific conditions, enabling the description of their constituent features and the ramifications of hyperparameter adjustments. 

In this paper, we introduce a novel approach for obtaining a dense vector capturing the contextual information of an individual word, employing the \ac{TM-AE} structure. Thereafter, we investigate the hyperparameters and demonstrate the influence of them on the distribution of the state space and \ac{RbE}. In light of the above issues, this study has the contributions below:

\begin{enumerate}
    \item Employing TM-AE to obtain a dense vector that captures the contextual information of an individual word.
    \item Investigating the hyperparameters to explore the state space of literals and \ac{RbE}.
    \item Enhancing classification performance through the utilisation of insights gained from the previous step.
\end{enumerate}

The remainder of the paper is organised as follows. Section~\ref{sec:related_work} presents the related work, while Section~\ref{sec:tsetlin_machine} summarises the operational concept of \ac{TM-AE} in this application.  Section~\ref{sec: proposed approach} provides a comprehensive description of the approach. In Section~\ref{sec:emprical_result}, the experiments are conducted for performance evaluation before we conclude the work in the last section.

\section{Background and Related Work}
\label{sec:related_work}
In the last two decades, with the profusion of computer resources, researchers were able to add more complexity to \ac{AI} algorithms, producing \ac{DL} in different domains~\cite{silver2013playing, he2016deep, goodfellow2014generative, radford2019language}.
For \ac{NLP}, the latest development in word embedding has made advanced strides~\cite{vaswani2017attention, devlin2018bert, hannun2014deep, van2016wavenet}. Firstly, a statistical \ac{NLP} replaced the symbolic \ac{NLP} to overcome the scalability issue~\cite{hutchins2004georgetown, weizenbaum1983eliza}. These \ac{NLP} models use statistical inference to learn the rules automatically~\cite{luhn1957statistical}. Later, using the \ac{NN} approach, Word2vec~\cite{mikolov2013efficient} and GloVe~\cite{pennington2014glove} were introduced to represent words as vectors in order to capture context and semantic similarity.

Recently, a promising \ac{ML} approach called the \ac{TM}~\cite{granmo2018tsetlin} has gained the researcher's attention. \ac{TM} has shown promising performance in terms of accuracy and memory utilisation~\cite{8798633,bhattarai2022word}. 
\ac{TM} paradigm requires less complexity, memory, and power footprint due to the linear combination of conjunctive clauses organised in lean parallel processing units. \ac{TM} has been utilised in NLP~\cite{yadav2021human, yadav2022robust}, where \ac{RbE} was often observed in \ac{TM} reasoning, indicated by the majority of negated features in clauses. For example, in~\cite{yadav2022robust},  the specificity parameter, \NspecificityS{}, is tuned to enhance negated reasoning during training, resulting in the generation of clauses that are robust to spurious correlations. 
\ac{RbE}  is not only applicable to \ac{NLP} but is also a fundamental cognitive process observed in human development. Infants demonstrate this ability through tasks that require them to reject certain options to deduce the correct one. For instance, in studies involving referent disambiguation, infants eliminate familiar objects when mapping novel labels to novel objects, effectively using a form of \ac{RbE} before they can even articulate logical components like ``or" and ``not" \cite{infantLogic,infantsRbE}. This parallel between \ac{AI} and human cognition underscores the relevance and applicability of \ac{RbE} across different domains.

For \ac{TM} embedding, it was developed in \cite{bhattarai2023tsetlin} a TM-based autoencoder that enables self-supervised learning of logical clauses, which outperformed its \ac{DL} counterparts, including GloVe~\cite{pennington2014glove}, achieving a high rating. The \ac{TM-AE}'s concept revolves around coalescing knowledge obtained from models supporting or not supporting the input tokens in a sequential manner over multiple stages until conjunctive clauses representing the resulting coalescence are generated. However, during this process, the model aims to generate clauses that provide a general description of the input sequence, making it challenging to extract the embedding for each token separately. 

This paper proposes a novel approach utilising the \ac{TM-AE} model to generate context information separately for each token in the vocabulary. This approach is similar to the \ac{DL} counterpart, Word2Vec, and specifically utilises the Skip-gram model, where the input is a token, and the output is a set of clauses describing the context in a dense vector. This approach enables the description of the space of states and the distribution of features concerning the input token through the generated clauses. Additionally, to improve training, we introduced the possibility of performing negated reasoning, named \ac{RbE}, utilising the margin parameter \NmarginT{} along with the specificity parameter\footnote{Different from \cite{yadav2022robust}, here we adopt the TM implementation with Boosting of True Positive Feedback~\cite{granmo2018tsetlin}, where $s=1$ can be configured. In \cite{yadav2022robust}, the vanilla TM implementation is employed, where $s>1$ always holds.  } \NspecificityS{}. The results indicate that \ac{RbE} provides a greater extent of control in enhancing model performance.
\section{Tsetlin Machine Autoencoder Basics}
\label{sec:tsetlin_machine}
\begin{figure}
    \centering
    \includegraphics[width=1.02\linewidth]{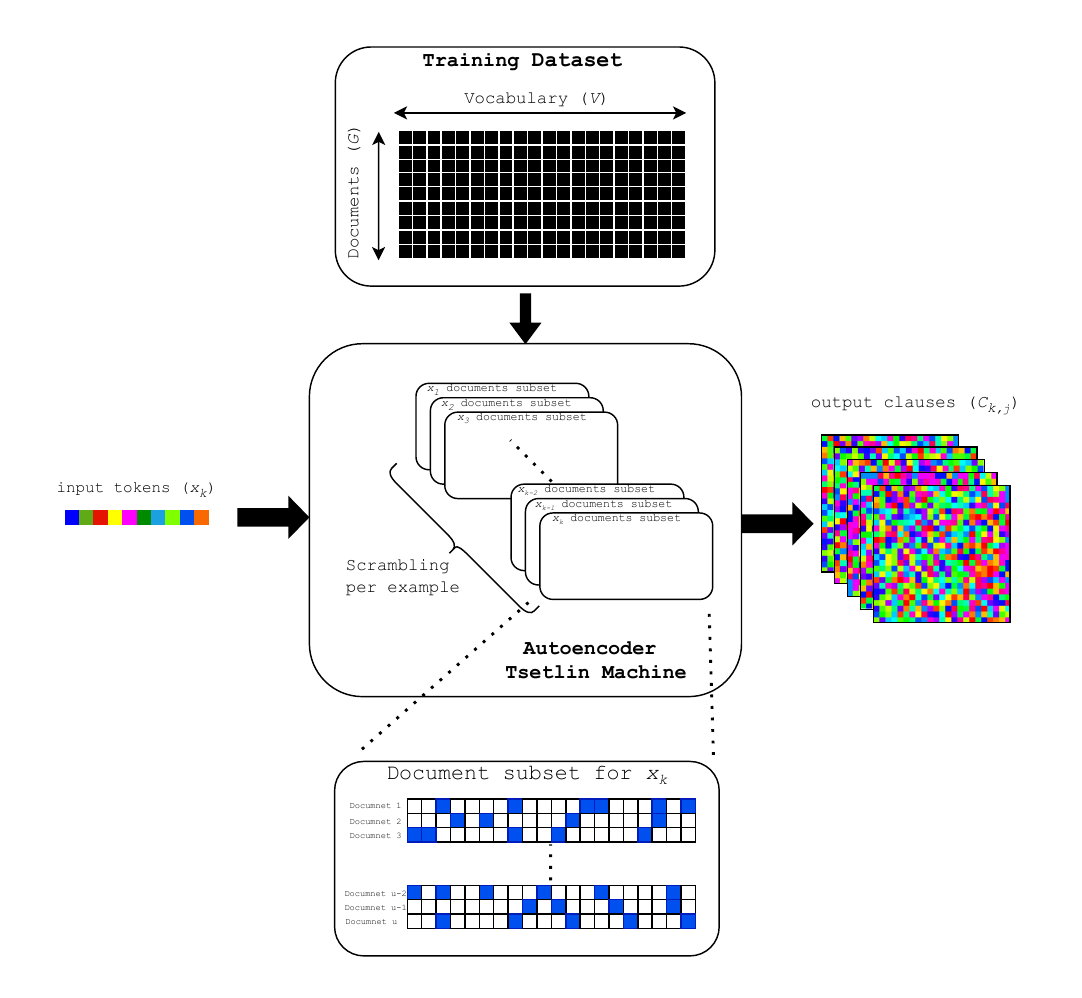}
    \caption{\ac{TM-AE} Architecture.}
    \label{fig:autoencoder_embedding}
\end{figure}

Here we review the basics of the \ac{TM-AE}~\cite{bhattarai2023tsetlin}, which was constructed based on the \ac{CoTM} algorithm. Its main objective was to generate embeddings for a sequence of input tokens ($x_1$, $x_2$, ..., $x_k$,..., $x_K$), where $k\in\{1,..., K\}$ is the index of tokens and $K$ represents the total number of desired selective tokens from the vocabulary $V$ to be embedded.
Within the \ac{CoTM} algorithm, the resulting embedding for each token $x_k$, is expressed using a collection of conjunctive clauses denoted as $C_{k,j}$, where $j$ varies from 1 to the user-defined number of clauses, $n$ (refer to Figure \ref{fig:autoencoder_embedding}).

Starting from this section, we will refer to the target input token, for which we aim to find the corresponding embedding, as the \ac{TW}. On the other hand, the tokens in the vocabulary that actively participate in the training process and contribute to the formation of clauses are referred to as features. The input to the \ac{TM-AE} model comprises a sparse matrix that represents the documents in the dataset as rows and the features in the vocabulary $V$ as columns. The model captures the vectorised representation of both the vocabulary $V$ and the documents $G$. During each epoch and with each iteration of examples, features in a document are selected by determining whether the document contains the \ac{TW} or not, based on the target value being either 1 or 0 respectively. The chosen documents corresponding to a specific \ac{TW} are combined with other documents for different \ac{TW}s using random shuffling during each example in an epoch.

\begin{figure}
    \centering
    \includegraphics[width=1.05\linewidth]{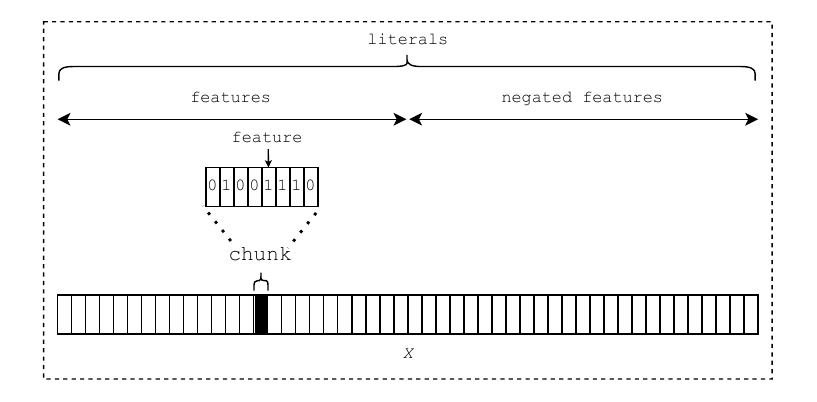}
    \caption{\ac{TM-AE} $X$-vector representation is stored in memory as chunks, where each chunk represents a specific block or group of related literals. Each chunk is implemented as a datatype, such as int8 bits, to enable more efficient memory usage and manipulation of feature representation.}
    \label{fig:x_vector}
\end{figure}

The selected documents corresponding to each \ac{TW} are combined, and their features are encoded into a one-dimensional binary vector $X$, thus achieving computational efficiency and resource optimisation during training (see Figure \ref{fig:x_vector}). The encoding of $X$ is such that each feature in the vocabulary $V$ and its negation are represented by two literals (one for the feature and one for its negation), which in turn contribute to the update process for building the clauses.

The binary vector $X$ encodes the binary representation of the presence or absence of each feature in the vocabulary, with respect to the selected documents, for every \ac{TW}. Furthermore, it encodes the negation of each feature along with its presence or absence. This mechanism enables an efficient encoding and processing of the information necessary for constructing the conjunctive clauses within the \ac{TM-AE} architecture.

The \ac{TM-AE} algorithm has the following steps:
\begin{enumerate}
    \item Data preparation and vectorisation, which involves determining the size of the vocabulary used in training.
    \item Target output identification.
    \item Scrambling the inputs to generate a unique and efficient embedding series.
    \item Collection of features that support the specified \ac{TW} if the output is $1$, or do not support it if the output is $0$.
    \item \ac{TM} update.
\end{enumerate}

The process of voting and building a clause within a \ac{TM}  can be analogised to a logical election. Features within the vocabulary engage in voting processes to favour a particular class within a formed clause. Features can vote for a \ac{TW} through multiple clauses, with a predetermined number of clauses to be constructed. To balance the learning of clauses for capturing distinct sub-patterns for a \ac{TW}, the \ac{TM} introduces a voting margin \NmarginT{}. Hyperparameter \NmarginT{} specifies the maximum number of learnt clauses (votes) for a certain sub-pattern. 

The \ac{TM} uses three update operations to learn and adapt to new input \ac{TW}: memorisation, forgetting, and invalidation. \textbf{Memorization} strengthens patterns by increasing the state of literals that match the input \ac{TW} and decreasing the states of missing literals randomly. \textbf{Forgetfulness}, on the other hand, weakens patterns by decreasing the state of true literals somewhat randomly. The hyperparameter \NspecificityS{}, which is a user-set parameter, influences the rate at which truth values are forgotten. Increasing \NspecificityS{} makes the patterns finer, while decreasing it results in coarser patterns. The \textbf{invalidation} operation increases the states of all false truth literals that the clause ultimately rejects. More details on \ac{TM} memory operations can be found in reference~\cite{DBLP:journals/corr/abs-2108-07594}.

The transparency of the \ac{TM-AE} algorithm prompted an exploration into understanding the rationale behind clause formation for a series of input \ac{TW}s in the final embedding~\cite{bhattarai2023tsetlin}. However, the model's scalability posed a challenge, resulting in the final outcome potentially containing fused or diluted contextual information, thereby potentially compromising its explainability. To address this, in this paper, context information was accumulated for each individual \ac{TW} as feature-equipped clauses. This approach involved considering the weight of each clause generated for the \ac{TW}. In the next section, more details on our approach will be provided.
\section{Proposed Approach}
\label{sec: proposed approach}
In this section, we will first present the approach of extracting contextual information for an individual TW and then detail the concept of reasoning by elimination, which is explored alongside the state space. 

\subsection{Context Information Extraction}
\label{sec:knowledge}
For \ac{TM-AE},  context information can be collected for a single \ac{TW} during the training phase. For this purpose, let 
\begin{equation} 
V = \{feature_1, feature_2, \ldots, feature_m\}.
\label{eq:vocab} 
\end{equation}

\begin{algorithm}
\caption{\ac{TM-AE} single-word embedding}
\label{alg:embedding}
\begin{algorithmic}[1]
\Require{$TW$: Target Word; $V$: Vocabulary; $TW \in V$; $D$: Documents; $G$: All the documents; $D \in G$, $u$: Windows size; $n$: The total number of clauses; $T$: Margin; $s$: Specificity; $r$: Rounds}
\State TMCreate$(n,T,s)$
\For{$r$ rounds}
    \State $q \gets$ Select$(\{0,1\})$
    \Comment{Random target value.}
    \If{$q = 1$}
        \State $G_{TW} \gets \{D|TW \in D\}$
        \Comment{Documents with $TW$.}
    \Else
        \State $G_{TW} \gets \{D|TW \notin D\}$
        \Comment{Documents without $TW$.}
    \EndIf
    \State $S \gets$ SelectU$(G_{TW},u)$
    \Comment{Random subset of window size $u$}
    \State $U \gets \bigcup_{D \in S} D$
    \Comment{Union of selected documents.}
    \State $X \gets (literal_1,literal_2,\ldots,literal_{2m})$, 
    \Statex where $literal_i = 
        \begin{cases}
        1, & {(feature_i \in U , 0 \leq i \leq m)}
        \\
        0, & {(feature_i \notin U , 0 \leq i \leq m)}
        \\
        1, & {(feature_i \notin U , m+1 \leq i \leq 2m)}
        \\
        0, & {(feature_i \in U , m+1 \leq i \leq 2m)}
        \end{cases}$
    \State TMUpdate($X, q$)
    \Comment{Update \ac{TM-AE} by $X$, and target value $q$ .}
\EndFor
\State $C, W \gets$ TMGetState()
\Comment{Get clauses $C$ and their weights $W$.}
\State \textbf{Return} $C,W$
\end{algorithmic}
\end{algorithm}

Algorithm \ref{alg:embedding} outlines the steps for embedding the \ac{TW} from the vocabulary $V$ that contains $m$ features. Firstly, the embedding algorithm applies $r$ rounds to process different combinations of documents. In each round, a subset $D$ of documents that contain the TW is selected from the complete set of documents $G$, assigning an output value of $q=1$. Alternatively, if the desired output value is 0 in the round, a different subset of documents $D$ that do not contain the \ac{TW} is selected from $G$. The \ac{TM-AE} then selects a specified proportion of documents based on a window size $u$.
The features in the selected documents are combined to facilitate the training for the \ac{TW}, and a Boolean one-dimension vector $X$ is generated. Meanwhile, while building $X$, the negated features are also included by inverting the status of the features, which expands the size of the $X$ vector to $2 \times m$.  Consequently, the vocabulary size in the TM-AE model is effectively doubled since it now includes both the features themselves and their negations, referred to as literals $L$, as Eq.~(\ref{eq:literals_vocab}).
\begin{equation} 
L = \{literal_1, literal_2, \ldots, literal_{2m}\}.
\label{eq:literals_vocab} 
\end{equation}

The vector space representation for \ac{TW} generated by the model consists of clauses with positive and negative weights as explained in Eq.~(\ref{eq:clause}). Here, $C_j^p(TW)$ denotes the $j^{th}$ clause with polarity $p$ for the particular TW. Polarity $p$ is either $+$ or $-$. Clearly, $C_j^p(TW)$ is a logical conjunction of literals $l_k$.  $L_j^p$ is the set of literals $l_k$, which contains the included/memorized literals in clause $C_j^p(TW)$, and is a subset of $L$. The length of $L_j^p$ may vary for each clause $C_j^p(TW)$, making it suitable to represent a wide range of textual features in \ac{TW}.

\begin{equation} 
C_j^p(TW) = \bigwedge_{l_k\in L_j^p} l_k.  
\label{eq:clause} 
\end{equation}

For example, the clause $C^+_1 (TW) = \neg literal_1 \wedge literal_2$ belongs to \ac{TW}, has index 1, polarity +, and consists of the literals $L^+_1 = \{ \neg literal_1, literal_2\}$. Accordingly, the clause outputs 1 if $literal_1$ = 0 and $literal_2$ = 1, and 0 otherwise.

This process enables the creation of an informative vector space description for the \ac{TW}, thereby facilitating subsequent used in diverse \ac{NLP} applications. In the knowledge graph, positive clauses for the target word ``queen'' include ``castle'', ``throne'', ``lady'', ``birthday'', ``princess'', ``diamond'', ``medal'', ``pop'', ``beauty'', ``Philip'', ``west'', ``king's AND speech'', ``park'', and ``ceremony''.
Despite the \ac{TM}'s output being generated in the form of clauses, this approach may result in a limited representation of the relationships between the \ac{TW} and other features within the vocabulary. In contrast, models such as Word2Vec have demonstrated the ability to capture context information in the form of dense vector representations for words, effectively describing their interactions within a high-dimensional space.
In the subsequent section, we will delve into a methodology that enables the capture of context information for all other features in the vocabulary, thereby facilitating a more comprehensive understanding of their relationships with the \ac{TW}.

\subsection{State Space and Reasoning by Elimination }
\label{sec:state_space}

\begin{figure}
    \centering
    \includegraphics[width=1.05\linewidth]{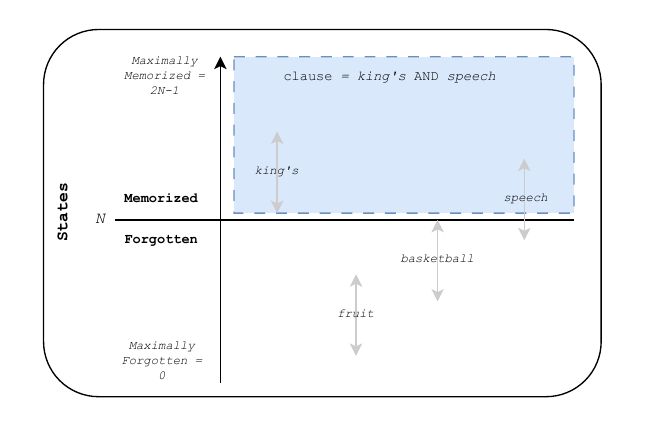}
    \caption{Single clause state space for the \ac{TW} ``queen". The blue area indicates the included content, i.e., (king's AND speech) in this example.}
    \label{fig:clause}
\end{figure}

To demonstrate the efficacy of the \ac{TM-AE} structure in capturing context information, we can consider the example of clauses captured by training the model for the \ac{TW} ``queen" which we mention in the previous subsection.
Each clause has a state space consisting of literals arranged in a distribution consistent with the context information captured during training (see Figure \ref{fig:clause}).

In the \ac{TA}\footnote{A \ac{TA} is the learning entity of a \ac{TM}. Please refer to \cite{granmo2018tsetlin} for more detail.} with $2N$ states, each state corresponds to a position that contributes to constructing a clause. The first $N$ states (ranging from 0 to $N-1$) are the forgotten states, in which the literal associated with these states does not actively participate in forming the final output of the \ac{TM} clause. In contrast, the region spanning from $N$ to $2N-1$ includes states where the literal is considered and contributes to the output as part of the clause (see blue area in Figure \ref{fig:clause}).
The distribution of literals in the state space is controlled by several hyperparameters, including \NspecificityS{}, \NmarginT{}, epochs, number of states, and window size. This behaviour determines the learning process and the included contextual information in clauses. 
This mechanism allows the \ac{TM} to dynamically adapt and prioritise relevant literals by reassigning their depth in memory and reinforcing their influence based on the voting mechanism. Consequently, the literals involved in the formation of clauses have states above $N$  and play a more prominent role. In contrast, less contributory literals are gradually pushed into the forgotten states.
This dynamic allocation of states and the selective concentration of literals contribute to the \ac{TW}'s ability to process and distil contextually significant information from large-scale datasets effectively.

The negations of features are affected in the same way but with a lower amount of pushing down to occupy states closer to involvement in clause formation. Through this approach, the model can increase the discrimination \ac{RbE}, and the model now is describing the \ac{TW} not only by what it looks like but also by what it does not look like, as exemplified by the Sherlock Holmes quote: 
\textit{When you have eliminated all which is impossible, then whatever remains, however improbable, must be the truth.}
 \ac{RbE} enables the \ac{TM} to generate AND-rules that embody the most critical words of a class that the \ac{TW} does not belong to, such as ``NOT fruit AND NOT basketball AND NOT ...," allowing for quick and robust learning without the need for randomising \ac{TM} feedback, which becomes deterministic. 

In the next section, we elucidate how insights gained from analysing the distribution of state spaces can enhance our understanding of \ac{TM}'s operation. We also show how hyper-parameter $s$ and $T$ influences \ac{RbE} and the performance of TM.

\section{Empirical Results}
\label{sec:emprical_result}
Empirical investigations were conducted utilising the \ac{TM-AE} with a single-clause embedding approach to analyse the distribution of the state spaces for distinct literals.
To this end, the One Billion Word dataset \cite{chelba2013one} comprising 30M samples and a vocabulary size of 40k words was employed. Other hyperparameter values were $N=2048$ and a window size $u$ of $25$. Contextual information was then extracted while manipulating various hyperparameters to observe their effects.

\begin{figure}
    \centering
    \includegraphics[width=1.0\linewidth]{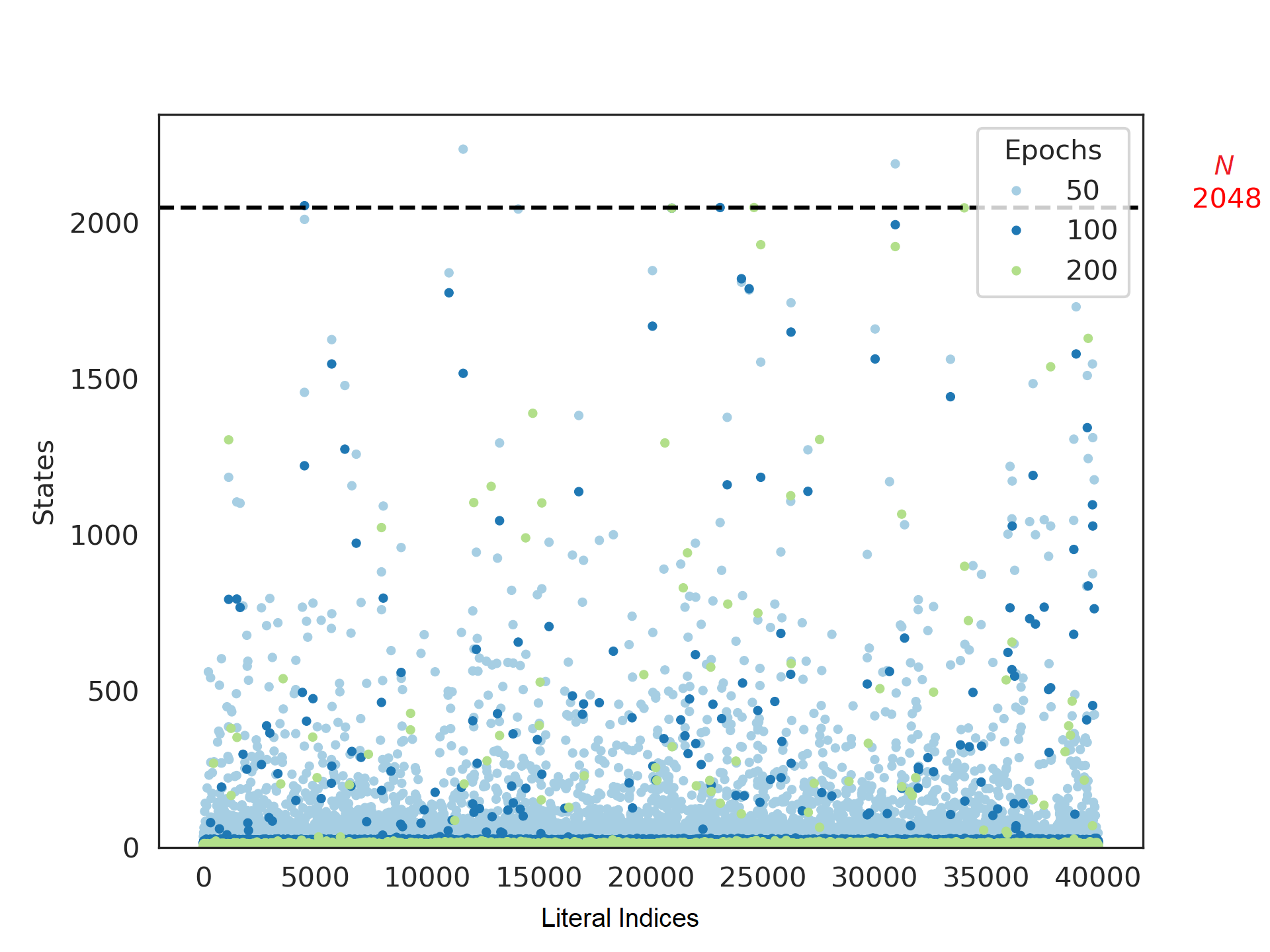}
    \caption{The Effect of Increasing Epochs on the Distribution of Literals in State Space for Original Features.}
    \label{fig:epochs_vs_space}
\end{figure}

One area of investigation involved examining the influence of increasing the number of epochs on the training process, with values tested at $5$, $100$, and $200$ epochs and other hyperparameter values were $T=3200$, $s=5.0$, as shown in Figure \ref{fig:epochs_vs_space}. The results demonstrated that as the number of iterations increased, the features progressively faded into oblivion, leading to the construction of concise clauses that effectively described the target class. This phenomenon is illustrated in the accompanying figure, where deeper levels of forgetting could be observed with higher repetition rates.

During initialisation, all literals are situated at the intermediate state in memory at $N$. Subsequently, updates are undertaken through three distinct mechanisms (as described in Section \ref{sec:tsetlin_machine}), with the distribution of literals determined by the influence of hyperparameters on the probability and speed of forgetting. In all figures, the dashed line labelled as $N$ appears at the top of the graphs, likely due to the fewer number of literals that exceed the $N$ threshold.

\begin{figure}
    \centering
    \includegraphics[width=1\linewidth]{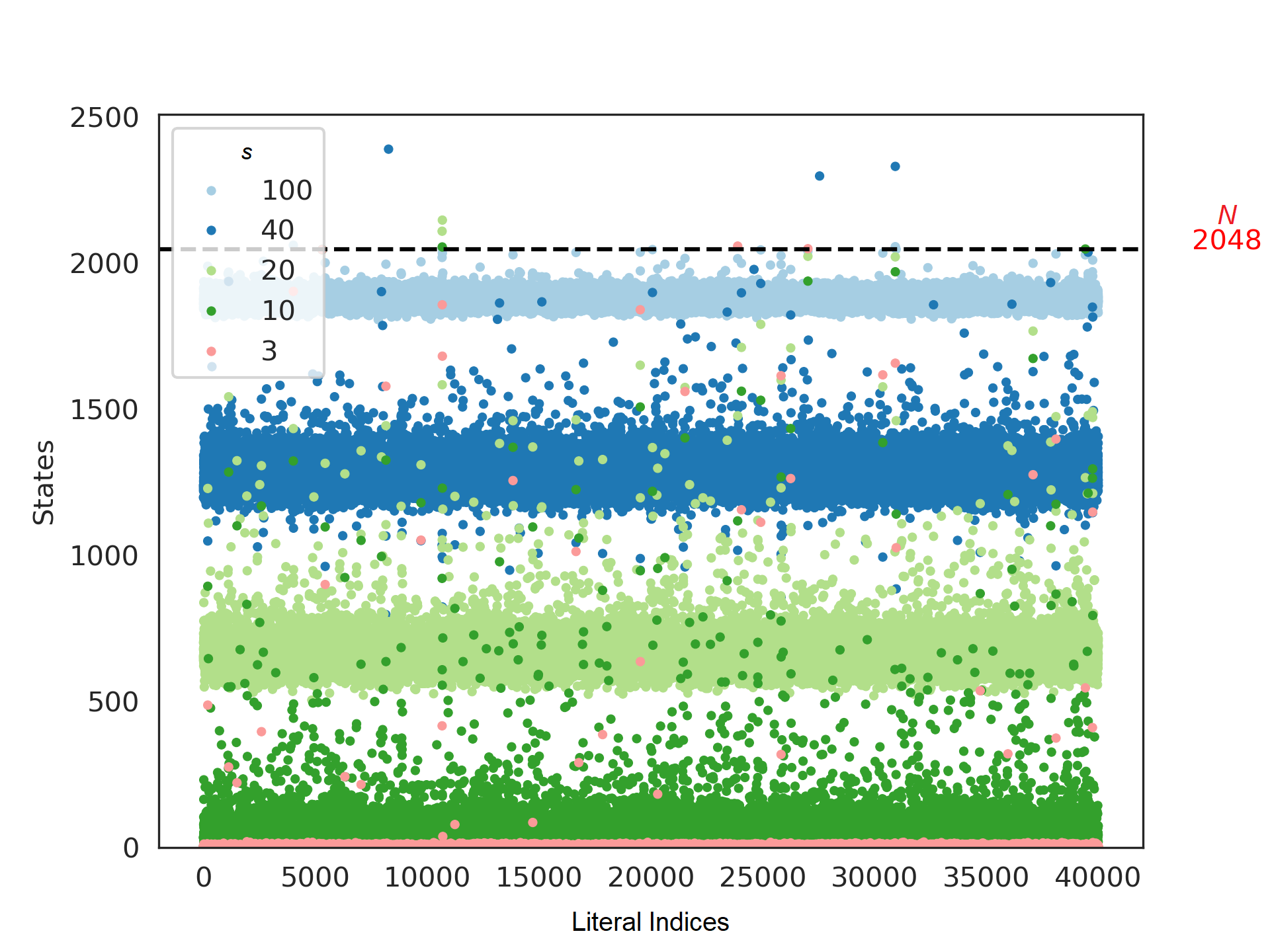}
    \caption{The Relationship between the Hyperparameter \NspecificityS{} and the Depth of Forgetting for Original Features.}
    \label{fig:s_vs_space}
\end{figure}

\begin{figure}
    \centering
    \includegraphics[width=1\linewidth]{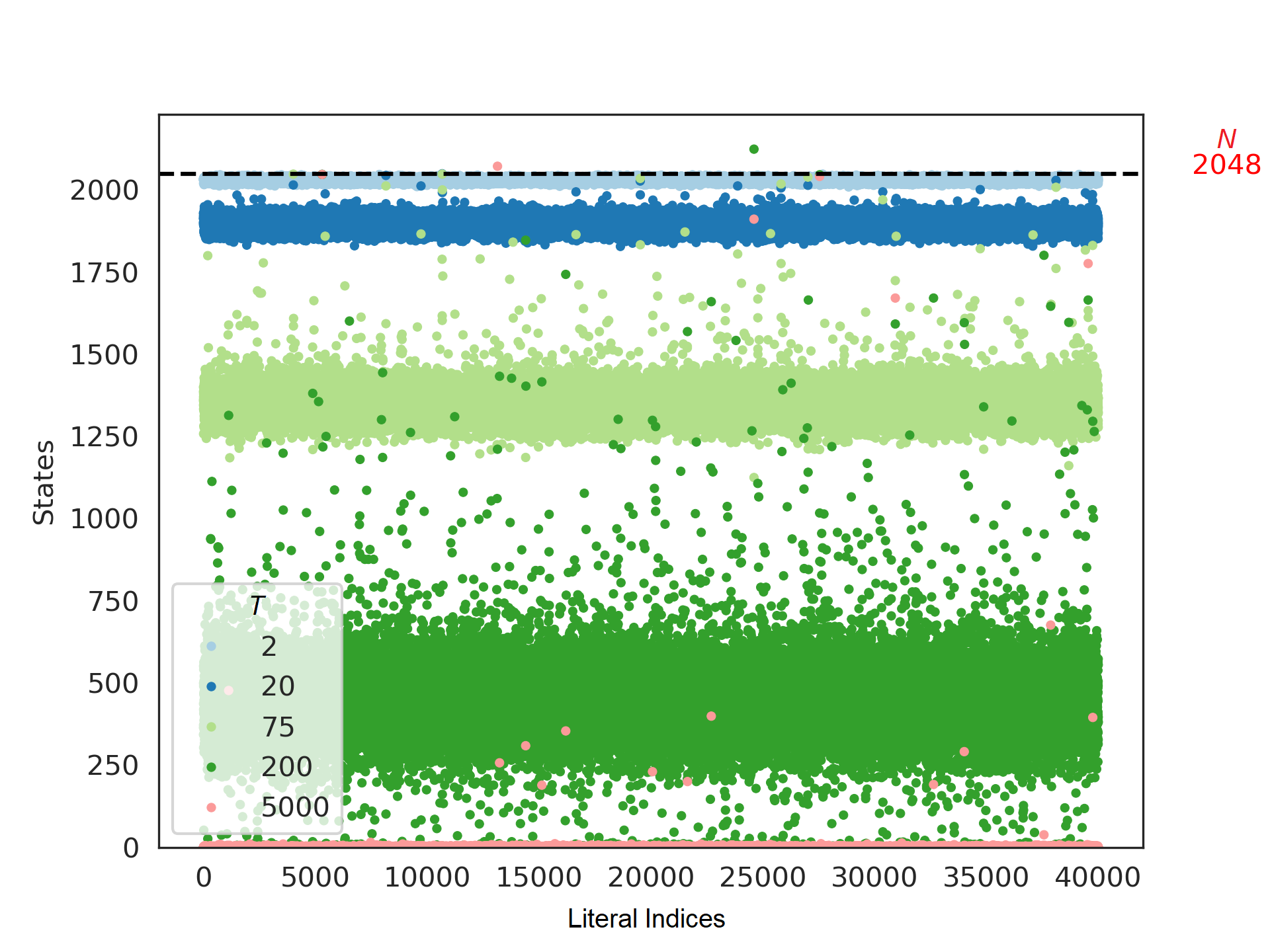}
    \caption{The Effect of Changing the \NmarginT{} Hyperparameter on the Distribution of Literals in State Space for Original Features.}
    \label{fig:T_vs_space}
\end{figure}

Several other experiments were conducted to investigate the impact of varying the \NmarginT{} and \NspecificityS{} hyperparameters on feature distribution in the state space with fixed 25 epochs. The results indicated that increasing the value of $s=$ (3, 10, 20, 40 and 100) reduced the depth of forgetting, thereby yielding more informed descriptions of the target class (as illustrated in Figure \ref{fig:s_vs_space}). 
While, as \NmarginT{} is increased (2, 20, 75, 200 and 5000) (Figure \ref{fig:T_vs_space}), the depth of forgetting increases, enabling a more discriminative representation of the \ac{TW}.


\begin{figure}
    \centering
    \includegraphics[width=1\linewidth]{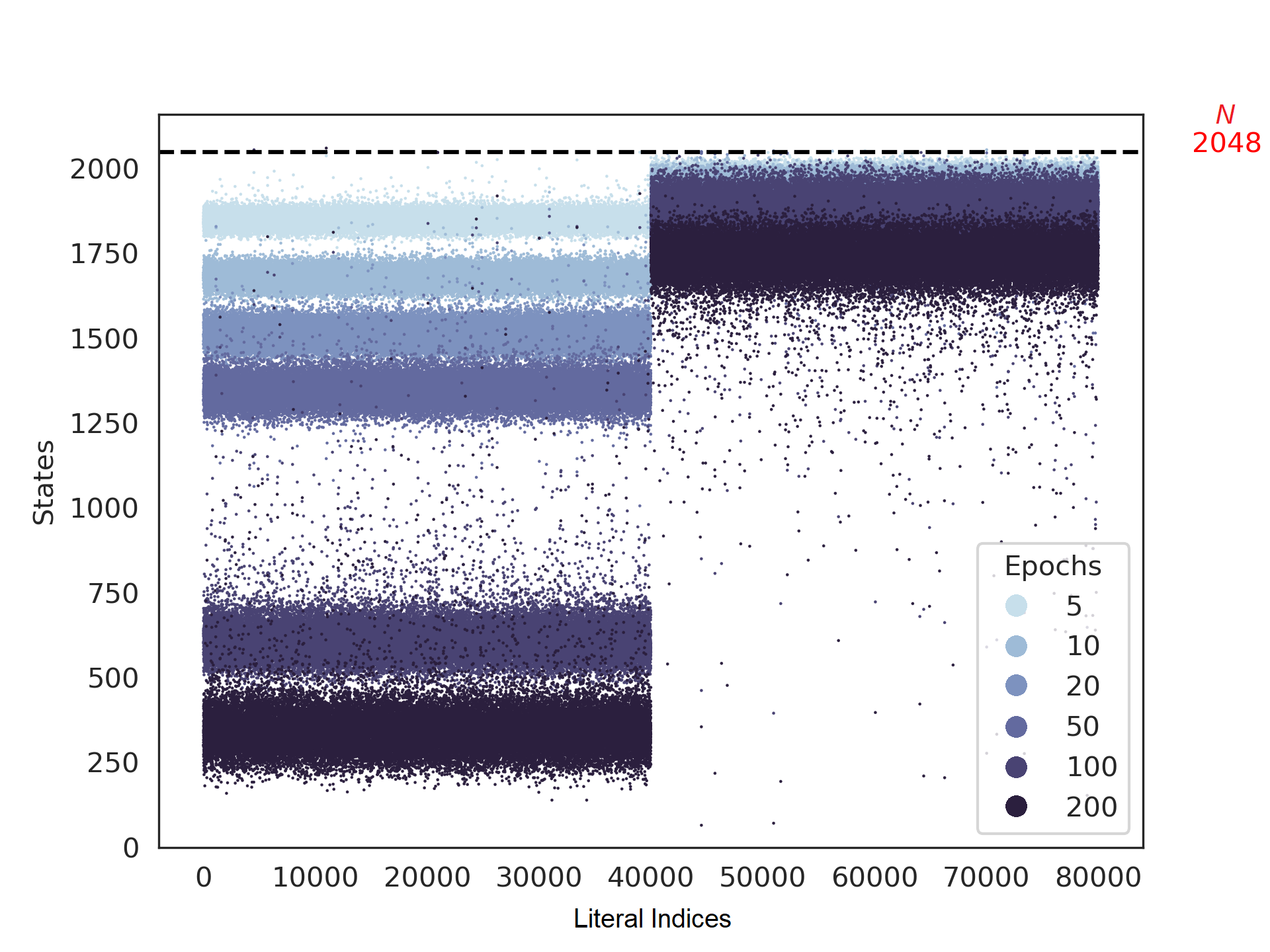}
    \caption{Impact of Epochs on Space Distribution for Original and Negated Features.}
    \label{fig:epochs_all}
\end{figure}

\begin{figure}
    \centering
    \includegraphics[width=1\linewidth]{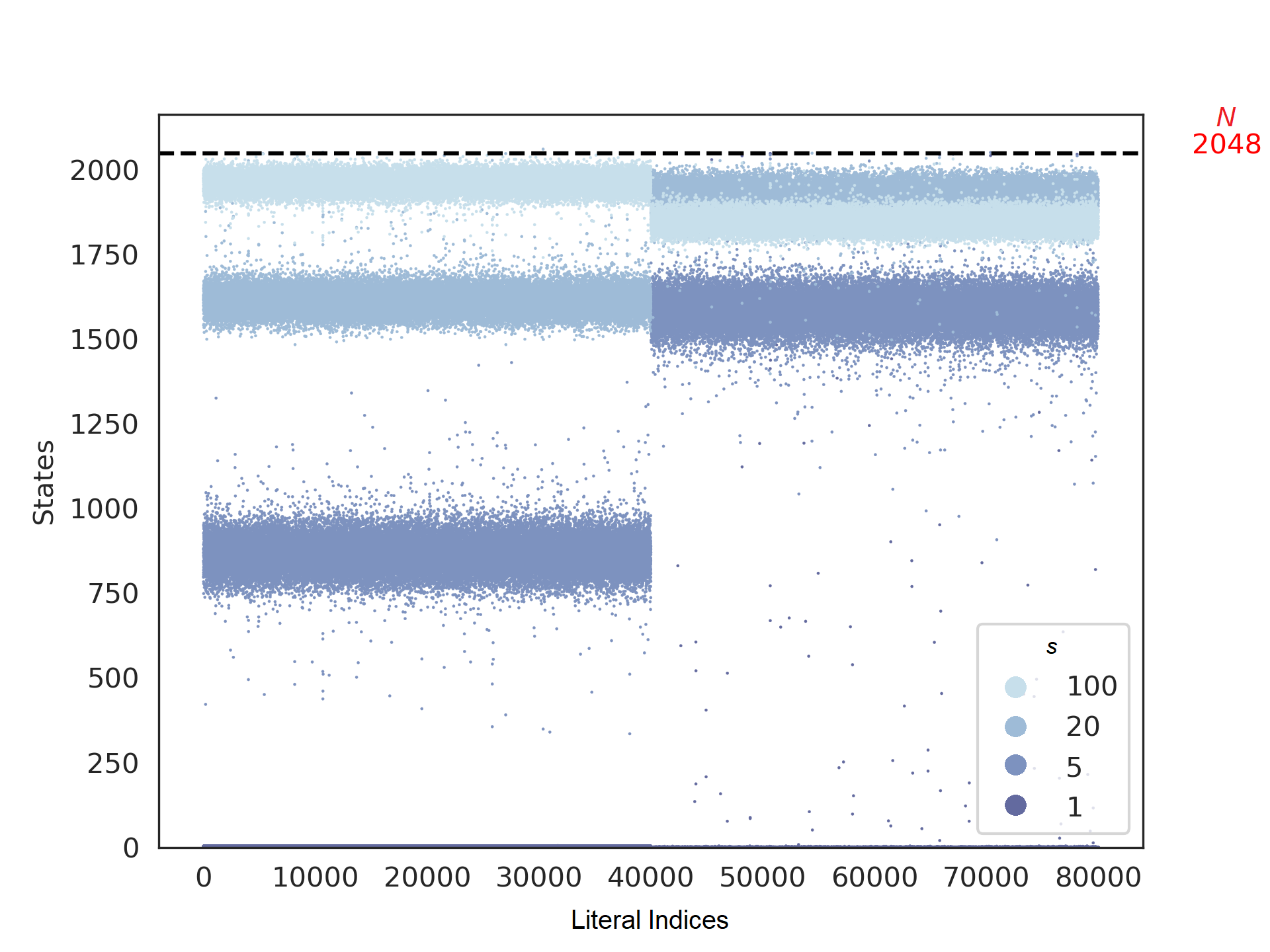}
    \caption{Effect of Hyperparameter \NspecificityS{} on State Distribution for Original and Negated Features.}
    \label{fig:s_all}
\end{figure}

Various experiments were conducted to explore the impact of feature negation, which entailed doubling the vocabulary size to form the $X$ vector. These experiments allowed for the observation of \ac{RbE}, particularly by comparing the state of literals before and after index 40000, which is the borderline between literals for the negation and the original form (Indices greater than 40000 represent negated ones, refer to Figure \ref{fig:x_vector} for x-axis). Generally, feature negation exhibited a slower speed of forgetting, resulting in their proximity to memorisation states and their potential contribution to clause formation.
The influence of increasing the number of epochs on the distribution of literals is depicted in Figure \ref{fig:epochs_all}. Meanwhile, Figure~\ref{fig:s_all} illustrates the effect of the hyperparameter \NspecificityS{} (specifically, values of 1, 5, 20, and 100) on the deepening of forgetfulness. A lower value of \NspecificityS{} signifies a greater tendency for the original features to be pushed to a deeper state and thus results in a reduced likelihood of their involvement in clause formation, whereas the negated features have shallower states. 
Furthermore, it is noteworthy that an increase in \NspecificityS{} induces a reversal in the positions of the original and negated features. For \NspecificityS{} values of 1, 5, and 20, the original features are pushed deeper compared to the negated features. However, at \NspecificityS{}=100, the negated features are found to occupy deeper states than the original features. This behaviour can be attributed to the specific implementation mechanics of the \ac{TM} and warrants further detailed exploration in future research endeavours.

\begin{figure}
    \centering
    \includegraphics[width=1\linewidth]{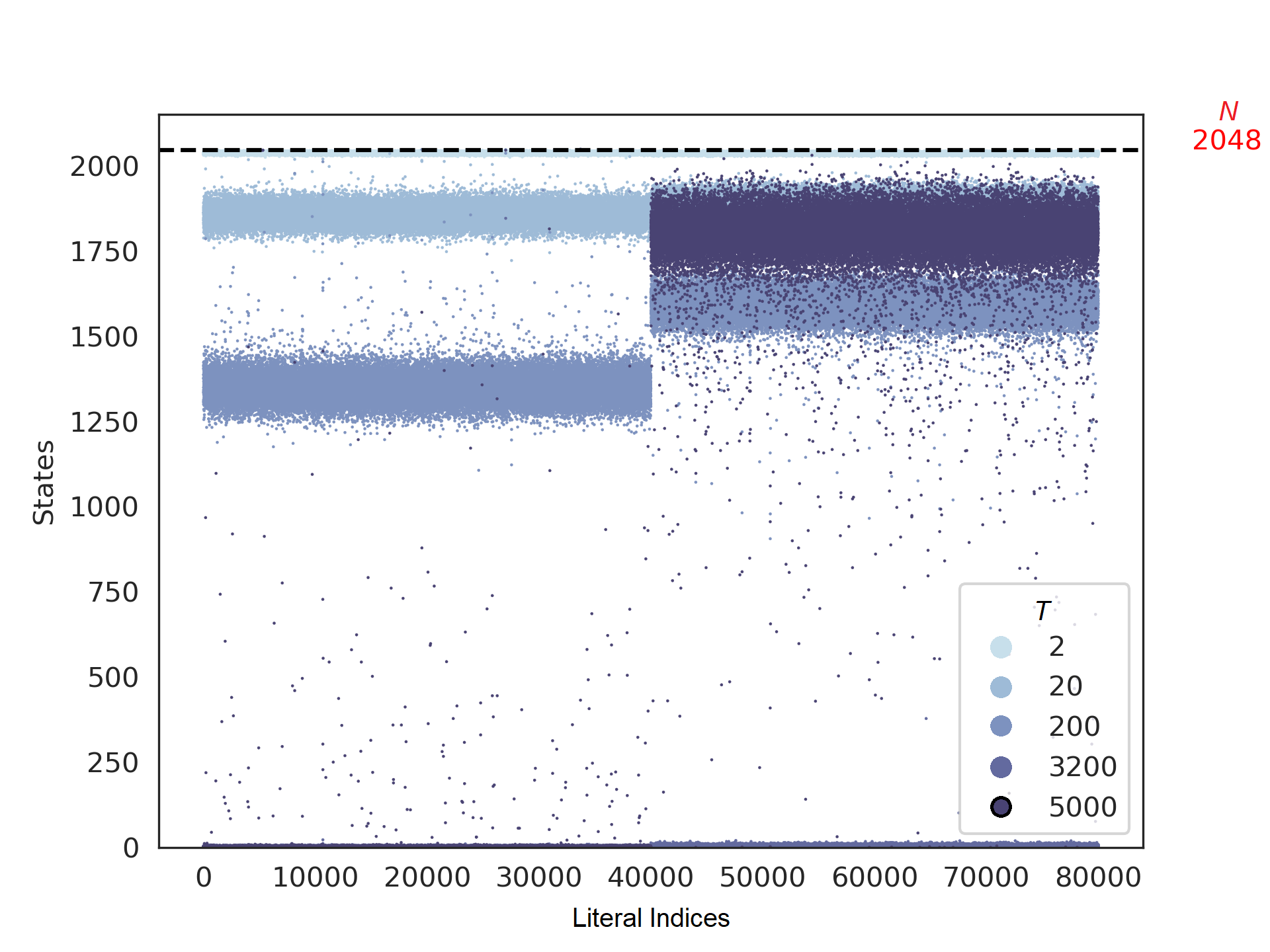}
    \caption{Impact of Hyperparameter \NmarginT{} on State Distribution for Original and Negated Features.}
    \label{fig:T_all}
\end{figure}

Figure \ref{fig:T_all} displays the experimental outcomes obtained by altering the hyperparameter \NmarginT{} (with values of 2, 20, 200, 3200, and 5000). Notably, \NmarginT{} exhibits an opposing effect to \NspecificityS{}, a larger margin leads to increased isolation of original features from participation and consequently allows for a greater involvement of negated features.
This observation underscores the critical role played by hyperparameters in shaping the embedding output of the \ac{TM}, highlighting the need to carefully select and fine-tune these hyperparameters to optimise the model's performance.


\begin{table*}[htbp]
    \centering
    \begin{tabular}{|c|c|c|c|c|c|c|c|c|} \hline 
         Dataset& $s=1$& $s=2$& $s=3$& $s=5$& $s=10$& $s=20$& $s=50$& Experiment setup\\ \hline 
         
         Artificial&  \textbf{60.88}&  49.38&  50.69&  50.62&  49.38&  51.22&  50.81& $T=10000,clauses=100$\\ \hline 
         
        IMDB& \textbf{90.62}& 86.23& 82.02& 79.65& 78.93& 78.4& 77.45& $T=10000,clauses=1000$\\ \hline 
        
        20 Newsgroups& \textbf{80.78}& 64.15& 54.54& 42.71& 40.13& 37.47& 31.33& $T=16000,clauses=160$\\ \hline 
    \end{tabular}
    \caption{Classification Accuracy of the \ac{TM} on Different Datasets for Various \NspecificityS{} Values.}
    \label{tab:s_change}
\end{table*}

\begin{table*}[htbp]
    \centering
    \begin{tabular}{|c|c|c|c|c|c|c|c|c|} \hline 
         Dataset& $T=4$& $T=8$& $T=16$& $T=32$& $T=64$& $T=128$& $T=256$& Experiment setup\\ \hline 
         
         Artificial&  51.95&  52.65&  53.18&  55.90&  57.7&  \textbf{58.74}& 55.85& $s=5.0,clauses=100$\\ \hline 
         
        IMDB& 55.095& 56.19& 62.72& 71.515& 75.38& 76.405& \textbf{77.56}& $s=5.0,clauses=100$\\ \hline 
 
        20 Newsgroups& 29.79& 30.88& 37.14& 48.60& 52.27& \textbf{52.71}& 50.86& $s=5.0,clauses=160$\\ \hline 
    \end{tabular}
    \caption{Classification Accuracy of the \ac{TM} on Different Datasets for Various \NmarginT{} Values.}
    \label{tab:T_change}
\end{table*}

Table \ref{tab:s_change} shows the experiment conducted to calculate the classification accuracy of the \ac{TM} on three different datasets: an artificial dataset, the IMDB dataset, and the 20 Newsgroups dataset. The artificial dataset was randomly generated with specific parameters, including the number of features, training examples, test examples, noise level, and unique characterising features per class. The IMDB dataset consists of 50k users' reviews for movies, while the 20 Newsgroups dataset contains news articles from various categories.

The classification accuracy in Table \ref{tab:s_change} was evaluated for different values of the parameter \NspecificityS{} (1, 2, 3, 5, 10, 20, and 50) to observe the performance variation. The results showed that for the artificial dataset, the \ac{TM} achieved the highest accuracy of 60.88\% when \NspecificityS{} was set to 1. However, as \NspecificityS{} increased, the accuracy slightly decreased. For the IMDB dataset, the highest accuracy achieved was 90.62\% when \NspecificityS{} was set to 1, with a gradual decrease in accuracy as \NspecificityS{} increased. Similarly, for the 20 Newsgroups dataset, the highest accuracy was 80.78\% at \NspecificityS{} equal to 1, with a decline in accuracy as \NspecificityS{} increased.

In addition to investigating the impact of varying the parameter \NspecificityS{}, we also examined the classification accuracy of the \ac{TM} on the same datasets for different values of the parameter \NmarginT{}. The results are summarised in Table \ref{tab:T_change}. For the artificial dataset, we observed that as \NmarginT{} increased, the accuracy initially improved until $T=128$, where the \ac{TM} achieved the highest accuracy of 58.74\%. 
Similarly, for the IMDB dataset, the highest accuracy of 77.56\% was achieved at $T=256$. For the 20 Newsgroups dataset, the best performance was 52.71\% and observed at $T=128$, surpassing the other \NmarginT{} values.

\begin{table}[htbp]
\caption{Effect of \ac{RbE} on the Number of Original and Negated Features for the IMDB Dataset.}
\label{tab:RbE}
\begin{tabular}{lccc}
\toprule
\textbf{Setup ($s,T$)} & 
\textbf{Original Feature} & 
\textbf{Negated Feature} & 
\textbf{Accuracy}\\
\midrule
(1, 256) & 0 & 1578 & 89.43\%\\
(3, 128) & 36 & 731 & 79.48\%\\
(10, 32) & 56 & 6 & 70.31\%\\
(50, 8) & 59 & 0 & 55.21\%\\
\bottomrule
\end{tabular}
\end{table}

The empirical results demonstrate that reinforcing \ac{RbE} with \NspecificityS{} to lower values (specifically 1) and \NmarginT{} to higher values can improve the accuracy of the \ac{TM}. Table \ref{tab:RbE} shows an experiment aimed at assessing the effect of \ac{RbE} for four configurations of \NspecificityS{} and \NmarginT{}. We extracted the classification accuracy of the \ac{TM} and counted the number of original and negated features present in 25 epochs of 50 positive weighted clauses. By encouraging \ac{RbE}, the \ac{TM} can push negated features to participate in forming the clauses. This process improves the selection of relevant features, which ultimately leads to better prediction outcomes.

\section{Conclusions}
\label{sec:conclusion}
In this paper, we employ the \ac{TM-AE} architecture to generate contextual information for individual words, facilitating the extraction of valuable insights. Building on this foundation, we investigate the \ac{RbE} configuration and the construction of the state space. We demonstrate the importance of efficiently distributing literals within the \ac{TM}'s clauses to enhance representation quality and capture essential contextual information. 
The empirical results show that integrating \ac{RbE} with lower values for $s$ and higher values for $T$ can enhance the accuracy of the \ac{TM}. By utilising \ac{RbE}, the \ac{TM} can direct the negation of features towards greater chances of taking part in forming the clauses and occupying the memorisation states. This procedure enhances the selection of relevant features, ultimately leading to improved predictive results.

\nocite{*}
\bibliographystyle{IEEEtran}
\bibliography{References}

\end{document}